\documentclass[10pt, a4paper]{article}

\usepackage[final]{lrec2026} % this is the new style
\usepackage{tabularx}
\usepackage{array}
\newcolumntype{Y}{>{\centering\arraybackslash}X}

\title{FLEURS-Kobani: Extending the FLEURS Dataset for Northern Kurdish}

\name{Daban Q. Jaff$^{1, 2}$, Mohammad Mohammadamini$^3$} 

\address{
  $^1$Erfurt University, Erfurt, Germany \\
  $^2$Koya University, Koysinjaq, Iraq \\
  $^3$LIUM, Le Mans University, Le Mans, France \\
  daban.hamad\_ameen@uni-erfurt.de, mohammad.mohammadamini@univ-lemans.fr
}

\abstract{
FLEURS offers n-way parallel speech for 100+ languages, but Northern Kurdish is not one of them, which limits benchmarking for automatic speech recognition and speech translation tasks in this language. We present FLEURS-Kobani, a Northern Kurdish (ISO 639-3 KMR) spoken extension of the FLEURS benchmark. The FLEURS-Kobani dataset consists of 5,162 validated utterances, totaling 18 hours and 24 minutes. The data were recorded by 31 native speakers. It extends benchmark coverage to an under-resourced Kurdish variety. As baselines, we fine-tuned Whisper v3-large for ASR and E2E S2TT. A two-stage fine-tuning strategy (Common Voice→FLEURS-Kobani) yields the best ASR performance (WER 28.11, CER 9.84 on test). For E2E S2TT (KMR→EN), Whisper achieves 8.68 BLEU on test; we additionally report pivot-derived targets and a cascaded S2TT setup. FLEURS-Kobani provides the first public Northern Kurdish benchmark for evaluation of ASR, S2TT and S2ST tasks. The dataset is publicly released for research use under a CC BY 4.0 license.
 \\ \newline  }

\begin{document}

\maketitleabstract

\section{Introduction}
Expanding available speech resources for new languages can significantly reduce the effort required for data curation across multiple tasks. Among widely used multi-task speech resources, the FLEURS \cite{Conneau2022} dataset supports few-shot learning and serves as a benchmark for several speech technologies, including automatic speech recognition (ASR), speech-to-text translation (S2TT), and Speech-to-Speech Translation (S2ST), across more than 100 languages. 
FLEURS  is a multilingual speech extension of FLORES \cite{Goyal2021} that provides spoken data for 101 of the 200 languages available in FLORES. Consequently, around 100 languages do not have corresponding speech data in FLEURS. Northern Kurdish is one of these languages. In this paper, we introduce the FLEURS-Kobani dataset, a spoken extension of the FLORES portion for Northern Kurdish.

A major obstacle in low-resource and multilingual NLP is not only training data scarcity, but also the lack of high-quality, standardized evaluation benchmarks that enable fair comparison across languages and research groups. FLORES-101 \cite{Goyal2021} was introduced to address this problem for machine translation (MT) by providing a carefully constructed benchmark: 3,001 sentences sampled from English Wikipedia and professionally translated into 200 languages under a controlled process, with broad topical coverage \cite{nllb2022}.

Building on this idea, FLEURS \cite{Conneau2022} extends the FLORES evaluation benchmark by adding the speech modality. FLEURS is an n-way parallel speech benchmark created by recording speech for sentences drawn from FLORES, yielding roughly $\approx$12 hours of speech per language. FLEURS is designed not as a single-task corpus but as a benchmark dataset for few-shot learning that can be used for multiple speech processing tasks, including automatic speech recognition (ASR), Spoken  Language Identification, Speech-to-Speech Translation (S2ST) and Speech-to-Text-Translation (S2TT) tasks \cite{Conneau2022}.

However, while the FLEURS benchmark includes Central Kurdish (ISO 639-3 CKB), it does not include Northern Kurdish, leaving it underrepresented in widely used benchmarking suites. Motivated by this gap and by the benchmark-extension approach exemplified by FLEURS, we introduce FLEURS-Kobani, a Northern Kurdish (KMR) multitask dataset built by recording 2,000 unique sentences from the Northern Kurdish portion of FLORES. The dataset comprises 5,162 utterances totaling 18 hours and 24 minutes of speech. To the best of our knowledge, this dataset is the first attempt to curate speech translation resources for Northern Kurdish. 

\subsection{Northern Kurdish}

Kurdish is an Indo-European language spoken across the Kurdistan region (Iraq, Iran, Turkey, and Syria) as well as by a sizable population in the diaspora. The number of native speakers is commonly reported to be over 30 million \cite{Opengin2021}. Its broad geographic dispersion and socio-political history have contributed to strong internal diversity, and Kurdish is often described as a dialect continuum \cite{Eppler2017} or a macrolanguage \cite{Sheyholislami2015}. 

Dialectological descriptions commonly distinguish major groupings such as Northern Kurdish (KMR), Central Kurdish (CKB), Southern Kurdish (SDH), Zazaki and Hewrami \cite{Haig2014, Mohammadirad2026}. Kurdish diversity is further reflected in writing practices shaped by geopolitics: Latin-based orthography is widely used for Northern Kurdish and Zazaki, whereas Arabic-based scripts are typical for Central Kurdish, Southern Kurdish and Hewrami groups \cite{Sheyholislami2015, Hassani2017}.

The focus of this study is Northern Kurdish, which is spoken in Turkey, Syria as well as parts of Iraq and Iranian Kurdistan, with additional speaker communities in countries such as Armenia, Lebanon and the east of Iran (Khorasan) \cite{Hassanpour1992, Izady1992}. Although Northern Kurdish is spoken by the majority of Kurdish speakers, it is often restricted or prohibited as a medium of education in Iran, Turkey, and Syria where the majority of speakers are situated \cite{BourgeoisGironde2026}. 

\subsection{Related Work}
In recent years, there have been significant efforts to provide speech resources for Kurdish dialects. In \cite{Veisi2022}, a phonetically balanced ASR dataset is designed for Central Kurdish. In \cite{Ahmadi2024}, an ASR dataset for Central Kurdish sub-dialects derived from TV series and comedy programs is developed. Common Voice 18 includes three Kurdish dialects, providing 68 hours of validated Northern Kurdish speech, 134 hours of Central Kurdish, and 1 hour of Zazaki \cite{commonvoice:2020}. In \cite{Hameed2025}, a collection of ASR resources for low-resource Middle Eastern languages is presented, which includes Laki, Hawrami, and Southern Kurdish branches of Kurdish.

To combat the resource scarcity in the speech translation domain, recent work has explored  Kurdish S2TT via pseudo-labeling, using an ASR+MT pipeline to create large pseudo-labeled Kurdish–English data for training end-to-end models \cite{Mohammadamini2025b}. Kuvost is a large-scale English to Central Kurdish S2TT dataset provided by the translation of English common Voice to Kurdish \cite{Mohammadamini2025a}. The FLEURS dataset \cite{Conneau2022} includes Central Kurdish but not Northern Kurdish.

Taken together, prior Kurdish speech research is unevenly distributed across dialects and tasks. The most substantial ASR and S2TT resources are concentrated on Central Kurdish. While Northern Kurdish is occasionally included in multilingual asr efforts \cite{omnilingualasr2025}, it remains largely missing from standard evaluation benchmarks for ASR, S2TT and S2ST tasks.

\section{Data Recording and Validation}

\subsection{Recording}
Data were recorded over a two-month period, from late April 2025 to late June 2025 at the University of Kobani via the Jiridastkrd\footnote{jiridastkrd.com} platform. A total of 31 speakers participated in the recordings: consisting of 26 female and 5 male speakers. Data collection followed three main phases: recruitment, outreach, and training.

\textbf{Recruitment and team setup:} In mid-April 2025, third-year students from the English Department, Faculty of Education were introduced to the project and invited to a two-hour orientation on procedures for recording Northern Kurdish sentences using the Jiridastkrd.com platform. Participants were informed that participation was voluntary, that they could withdraw at any time, and that they could request deletion of their recordings without needing to provide a reason. Of the nine students initially contacted, eight continued with the project.

\textbf{Outreach and speaker diversification:} Because the initial student group was small and entirely female, the eight participants were asked to help recruit additional speakers, especially male speakers, to improve gender and speaker coverage.

\textbf{Training and recording procedure:} Students were trained on (i) reading the prompt before recording, (ii) how to begin reading only after pressing the record button, and (iii) how to upload recordings via the platform. Additional recruits received the same orientation.

\subsection{Validation}
Validation was conducted within the Jiridastkrd.com workflow by reviewing each recording together with its associated Northern Kurdish prompt and metadata. We applied two classes of checks: prompt conformity (whether the speaker read the intended sentence) and audio quality (whether the recording is sufficiently intelligible). Recordings were flagged for exclusion if they met any of the following criteria:

\begin{itemize}
    \item[(i)] \textbf{Prompt mismatch/misreading:} the spoken content does not match the assigned Northern Kurdish prompt (e.g., reading a different sentence or substantial divergence from the prompt).
    \item[(ii)] \textbf{Truncation:} a word (or multiple words) is cut off at the beginning, middle, or end of the recording.
     \item[(iii)] \textbf{Insertions:} the speaker adds extra word(s) not present in the prompt (beyond minor filled pauses). 
     
    \item[(iv)] \textbf{Pronunciation errors:} the pronunciation renders a word difficult to recognize or changes the intended lexical item; as exceptions, we did not enforce a single ``correct'' reading for foreign proper names and abbreviations, which appear in the FLEURS prompts and were read inconsistently across speakers.
    \item[(v)] \textbf{Excessive background noise:} or overlapping speech that reduces intelligibility.
    \item[(vi)] \textbf{Very low recording level:} (low volume) such that the speech cannot be heard clearly under normal listening conditions.
\end{itemize}

Socio-political and technical constraints in the Rojava region directly influenced the recording process. A primary limitation is the significant demographic imbalance; the project was initiated with a core team of eight female students, and despite subsequent outreach efforts, the team remains heavily skewed, with 26 female speakers compared with only 5 male participants. Data collection was further hampered by unstable internet connectivity and limited digital infrastructure outside of the University of Kobani, and the reliance on personal cellphones for recordings contributed to a high rate of invalid entries due to varying microphone quality and technical inconsistencies. 

Furthermore, many participants reported limited formal education in Kurdish, with most having started studying it only in seventh grade as an elective subject. This negatively affected reading fluency, increased variability in the recordings, and resulted in specific linguistic inconsistencies, particularly in the pronunciation of abbreviations and foreign proper names, including the names of cities and countries. Although direct prompt mismatches were relatively limited, other technical errors were very common, including truncation in which words were cut off, excessive background noise, and pronunciation errors that rendered words difficult to recognize. Additionally, a significant number of recordings were excluded due to very low recording levels that made speech unintelligible. As a result, of the 7,897 Northern Kurdish recordings, 5,162 were marked as good quality (65.5\%), while 2,735 were marked as low quality (34.5\%).

\section{Data Specifications}
The resulting validated data comprise 5,162 utterances, equivalent to 18 hours and 24 minutes of speech. The validated utterances are partitioned into three subsets, according to the English version of the FLEURS dataset. For a given file, if its English translation matches a sentence in the English FLEURS dataset, it is assigned to the corresponding train/dev/test split. We found 149 samples that did not match any sentence in the English FLEURS dataset. These utterances are placed into the dev set in order to increase the number of files in this partition from 342 to 491 utterances. The numbers of utterances and the duration of each partition are presented in Table~\ref{tab:dataspecific}. The dataset can be accessed through a dedicated Hugging Face repository \footnote{\url{https://huggingface.co/datasets/aranemini/northern-kurdish-fleurs}}.

\begin{table}[h]
\centering
\small
\begin{tabularx}{\linewidth}{|l|Y|Y|Y|}
\hline
\textbf{Partition} & \textbf{Train} & \textbf{Dev} & \textbf{Test} \\
\hline
Utterances & 3,870 & 491 & 801 \\
Duration     & 13:47 & 1:42 & 2:55 \\
Speakers   & 30 & 20 & 26 \\
\hline
\end{tabularx}
\caption{FLEURS-Kobani benchmark specifications}
\label{tab:dataspecific}
\end{table}

\section{Experiments and Results}
We evaluated our dataset using the Whisper v3-large model for both ASR and S2TT tasks \cite{whisper}. The Whisper model was fine-tuned for five epochs with a learning rate initialized at 1e-5. Evaluation on the validation set was performed every 50 training steps. The batch size was set to 4 during training. The checkpoint achieving the best performance on the validation set was selected for evaluation on the test set.

\subsection{ASR Results}
The dataset was evaluated on the ASR task under three experimental conditions. First, we fine-tuned the Whisper model using the Northern Kurdish subset of Common Voice, which contains 68 hours of validated speech. This configuration achieved a WER of 36.28 on the development set and 37.02 on the test set. Second, we fine-tuned the Whisper v3-large model using the training portion of FLEURS-Kobani, obtaining a WER of 34.86 on the development set and 38.09 on the test set. Finally, we fine-tuned the Whisper model on FLEURS-Kobani after an initial fine-tuning stage on Common Voice. This two-stage fine-tuning strategy resulted in a WER of 28.53 on the development set and 28.11 on the test set. The results are reported in Table \ref{tab:asr}.

\begin{table}[h]
\centering
\small
\begin{tabular}{|l|l|l|l|}
\hline
\textbf{-} & \textbf{CV} & \textbf{F-K} & \textbf{CV  $\rightarrow$  F-K} \\
\hline
DEV & 12.33/36.28  & 13.98/34.86 & 10.11/28.53  \\
TEST & 12.28/37.02 &  15.97/38.09 & 9.84/28.11 \\

\hline
\end{tabular}
\caption{ASR results on KMR FLEURS benchmark (CER/WER), CV: Common Voice, F-K: FLEURS-Kobani}
\label{tab:asr}
\end{table}

\subsection{E2E S2TT Results}
As a first E2E S2TT experiment, we fine-tuned the Whisper v3-large model on the training portion of the FLEURS-Kobani dataset. This configuration achieved BLEU scores of 9.16 and 8.68 on the development and test sets, respectively.
For other languages available in the FLEURS dataset, English was used as a pivot language. Using this pivot, we created parallel KMR–FR, KMR–NL, KMR–CKB, and KMR–FA versions of the train/dev/test splits of the FLEURS-Kobani dataset. We observed that target languages such as NL and FR, which are closer to the pivot language, yield better results.
For the KMR–CKB and KMR–FA pairs, despite their linguistic similarity to KMR, we did not achieve significant results. This behavior may be attributed to divergences introduced by lexical and typological differences between the target languages and the pivot language. The E2E S2TT results are presented in Table \ref{tab:s2tt}.

\begin{table}[h]
\centering
\small
\begin{tabularx}{\linewidth}{|l|Y|Y|}
\hline
\textbf{Target} & \textbf{DEV} & \textbf{TEST} \\
\hline
EN  & 9.16/34.72  & 8.68/34.15 \\
FR  & 5.00/27.94  & 6.16/29.78 \\
NL  & 2.70/25.71  & 2.82/26.62 \\
CKB & 0.84/19.11  & 1.24/20.83 \\
FA  & 0.94/20.15  & 1.36/21.05 \\
\hline
\end{tabularx}
\caption{E2E S2TT results (BLEU/ChrF++) from Northern Kurdish (KMR) to five target languages/dialects.}
\label{tab:s2tt}
\end{table}

\subsection{Cascaded S2TT}
In the final experiment, we evaluated our dataset on the cascaded S2TT task. First, we transcribed the KMR audio using an ASR model (Whisper fine-tuned on CV and FLEURS-Kobani). Then, the ASR outputs were translated using NLLB-1.3B \cite{nllb2022}. The cascaded results are reported in the ASR+MT column of Table \ref{tab:cascaded}. The first column of Table \ref{tab:cascaded} presents the machine translation (MT) results obtained by translating the references, which serve as the baseline.

\begin{table}[h]
\centering
\small
\begin{tabularx}{\linewidth}{|l|Y|Y|}
\hline
\textbf{Split} & \textbf{MT (Base)} & \textbf{ASR+MT} \\
\hline
Dev  & 24.98/51.68 & 16.98/43.91 \\
Test & 27.36/52.96 & 19.60/45.51 \\
\hline
\end{tabularx}
\caption{Cascaded (KMR$\rightarrow$EN) S2TT results (BLEU/ChrF++): MT(Base) translation of references using NLLB; ASR+MT translates ASR outputs using the NLLB model.}
\label{tab:cascaded}
\end{table}

\section{Conclusion}
This paper introduced FLEURS-Kobani, a speech extension of FLEURS for Northern Kurdish (Kurmanji; KMR). The dataset helps address a key resource gap by enabling benchmarking of core speech technologies such as ASR, speech-to-text translation (S2TT), and Speech-to-Speech Translation (S2ST) for Northern Kurdish. By releasing FLEURS-Kobani along with baseline results, our aim is to lower the barrier for future research and accelerate progress on Kurdish speech technology, with benefits for accessibility and language inclusion. 
Since there are two variables—fluency and dialect—in the utterances, a more granular evaluation at the speaker level, as well as an analysis of regional dialect variability in comparison to the written form, will provide a clearer view of the obtained results in future research.
This work encourages future research to expand coverage to additional Kurdish varieties.

\section*{Acknowledgements}

The authors express their deepest gratitude to the following individuals for their dedicated support and contributions to the project:

\par
\mbox{\textbf{Arîvan}~Misî}, \mbox{\textbf{Bahoz}~Makû}, \mbox{\textbf{Bushra}~Ali}, \mbox{\textbf{Cedra}~Hami}, \mbox{\textbf{Diane}~Mohamed}, \mbox{\textbf{Dilava}~El~Ehmed}, \mbox{\textbf{Dilê}~Berkel}, \mbox{\textbf{Eslem}~Ahmed}, \mbox{\textbf{Falak}~Hitto}, \mbox{\textbf{Hacra}~Îsa}, \mbox{\textbf{Kaniwar}~Hisên}, \mbox{\textbf{Kemal}~Besrawî}, \mbox{\textbf{Mehmûd}~Elî}, \mbox{\textbf{Mehrîvan}~Hussain}, \mbox{\textbf{Mistefa}~Temo}, \mbox{\textbf{Narîn}~Bedir}, \mbox{\textbf{Nazdar}~Ebde}, \mbox{\textbf{Naze}~Afandi}, \mbox{\textbf{Nujeen}~Ebda}, \mbox{\textbf{Ranya}~Heyder}, \mbox{\textbf{Rojan}~Bozan}, \mbox{\textbf{Roz}~Mistefa}, \mbox{\textbf{Ruho}~Mihemed}, \mbox{\textbf{Salan}~Bozan}, \mbox{\textbf{Sara}~Hussain}, \mbox{\textbf{Serbest}~Bozan}, \mbox{\textbf{Shilan}~Afandi}, \mbox{\textbf{Silva}~Haj~Ali}, \mbox{\textbf{Talîn}~Mistefa}, \mbox{\textbf{Yasmin}~Mustefa}, \mbox{\textbf{Zehra}~Xelîl}.
\par

Our sincere thanks also go to \textbf{Kobani University} for facilitating the project and providing the necessary institutional support. Additionally, we would like to extend a special note of thanks to \textbf{Razaw Bor} for her invaluable assistance in validating portions of the dataset.  

Daban Q. Jaff gratefully acknowledges the support of the \textbf{Deutscher Akademischer Austauschdienst (DAAD)} through a PhD research grant (Grant No.~57645448) for his doctoral studies at the University of Erfurt (Host: \textbf{Language and Its Structure, Prof.~Dr.~Beate Hampe}). Mohammad Mohammadamini acknowledges the  \textbf{LIUM (Le Mans University)} for hosting and providing the computational resources used in the experiments. Finally, the authors thank the anonymous reviewers for their valuable comments.

\section{References}

\bibliographystyle{lrec2026-natbib}
\bibliography{lrec2026-example}

\end{document}